\documentclass[conference,a4paper]{IEEEtran}
\IEEEoverridecommandlockouts

\usepackage[hidelinks]{hyperref}
\usepackage[cmex10]{amsmath}
\usepackage{amssymb,amsfonts}
\usepackage{dblfloatfix}
\usepackage[ruled,vlined]{algorithm2e}
\usepackage{graphicx}
\graphicspath{{Figures/PDF/}{Figures/PNG/}}
\usepackage{booktabs}
\usepackage{siunitx}
\usepackage[numbers,compress]{natbib}
\usepackage{texnames}
\usepackage{bm,bbm}
\usepackage{orcidlink}
\usepackage{bbding}

\begin{document}

\title{
Low-Rank Adaptation of Pre-Trained Stable Diffusion for Rigid-Body Target ISAR Imaging 
\thanks{$\dagger$ denotes equal contribution. 
}
}

\author{
	\IEEEauthorblockN{Boan Zhang $\dagger$}
	\IEEEauthorblockA{\textit{Xidian University}\\
		Xi'an, China, 710126 \\
        boan@stu.xidian.edu.cn
		}
        \\
        \IEEEauthorblockN{Rongrong Wang}
	\IEEEauthorblockA{\textit{Xidian University}\\
		Xi'an, China, 710126\\
        wangrongrong@xidian.edu.cn
        }

        \and

        \IEEEauthorblockN{Hang Dong $\dagger$}
	\IEEEauthorblockA{\textit{Xidian University}\\
		Xi'an, China, 710126 \\
        hdong029@stu.xidian.edu.cn
        }
        
        \\
        \IEEEauthorblockN{Zhenhua Wu}
	\IEEEauthorblockA{\textit{Anhui University}\\
		Hefei, China, 230601 \\
        zhwu@ahu.edu.cn
        }
        
	\and
	\IEEEauthorblockN{Jiongge Zhang $\dagger$}
	\IEEEauthorblockA{\textit{Xidian University}\\
		Xi'an, China, 710126 \\
        jgzhang@stu.xidian.edu.cn
        }
        \\

        \IEEEauthorblockN{Xiyang Liu}
	\IEEEauthorblockA{\textit{Xidian University}\\
		Xi'an, China, 710126 \\
        xyliu@xidian.edu.cn
        }

        \and
        \IEEEauthorblockN{Long Tian \Envelope}
	\IEEEauthorblockA{\textit{Xidian University}\\
		Xi'an, China, 710126\\
        tianlong@xidian.edu.cn
        }
        \\

        \IEEEauthorblockN{Hongwei Liu}
	\IEEEauthorblockA{\textit{Xidian University}\\
		Xi'an, China, 710126 \\
        hwliu@xidian.edu.cn
        }
        
}

\maketitle

\begin{abstract}
Traditional range-instantaneous Doppler (RID) methods for rigid-body target imaging often suffer from low resolution due to the limitations of time-frequency analysis (TFA). To address this challenge, our primary focus is on obtaining high-resolution time-frequency representations (TFRs) from their low-resolution counterparts. Recognizing that the curve features of TFRs are a specific type of texture feature, we argue that pre-trained generative models such as Stable Diffusion (SD) are well-suited for enhancing TFRs, thanks to their powerful capability in capturing texture representations. Building on this insight, we propose a novel inverse synthetic aperture radar (ISAR) imaging method for rigid-body targets, leveraging the low-rank adaptation (LoRA) of a pre-trained SD model. Our approach adopts the basic structure and pre-trained parameters of SD-Turbo while incorporating additional linear operations for LoRA and adversarial training to achieve super-resolution and noise suppression.
Then we integrate LoRA-SD into the RID-based ISAR imaging, enabling sharply focused and denoised imaging with super-resolution capabilities.
We evaluate our method using both simulated and real radar data. The experimental results demonstrate the superiority of our approach in frequency estimation and ISAR imaging compared to traditional methods. Notably, the generalization capability is verified by training on simulated radar data and testing on measured radar data.
\end{abstract}


\section{Introduction}
Inverse synthetic aperture radar (ISAR) imaging has gained widespread use in both civilian and military applications due to its long-range, all-day, and all-weather observation capabilities \citep{Du2005, Cumming2004, Tian2020}. 
Under uniform target motion, the range Doppler (RD) algorithm \citep{Cumming1992, Skolnik1980, Zhang2010}, which involves translational motion compensation followed by conventional Fourier transform (FT), can effectively achieve high azimuthal resolution. However, in real-world applications, target motion is often complex and non-uniform. As a result, Doppler frequencies become time-varying, making them unsuitable for accurate estimation using the FT-based RD algorithm. 
Time-frequency analysis (TFA) methods have been developed for ISAR imaging to mitigate the blurred imaging caused by the time-varying motion of rigid-body targets, commonly referred to as range-instantaneous Doppler (RID) methods \cite{Chen2016}.


\begin{figure}[t]
\centering
\centerline{\includegraphics[width=9cm]{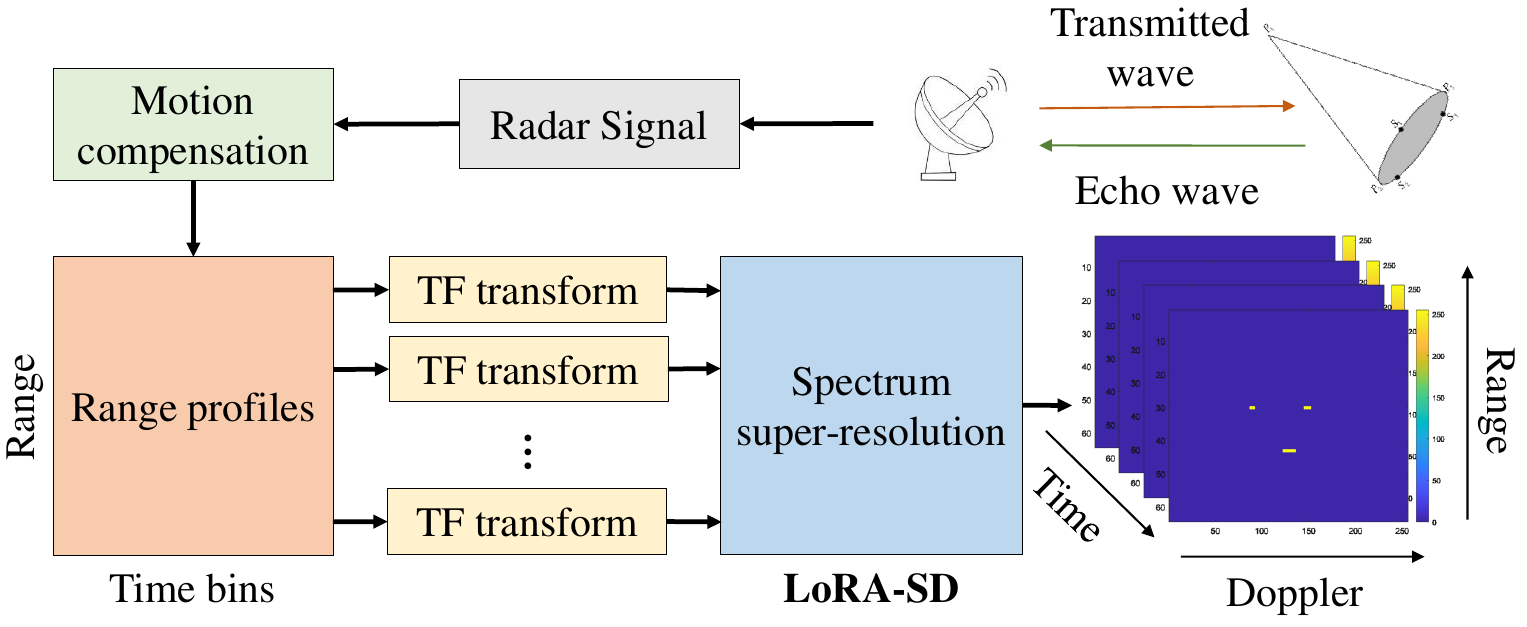}}
\caption{\small Flowchart of our proposed TFA-based ISAR imaging. In order to obtain high-resolution TFRs, we replace the original time sampling in RID with our LoRA-SD for spectrum super-resolution.} 
\label{fig:model}
\vspace{-3mm}
\end{figure}

\begin{figure*}[t]
\centering
\centerline{\includegraphics[width=17cm]{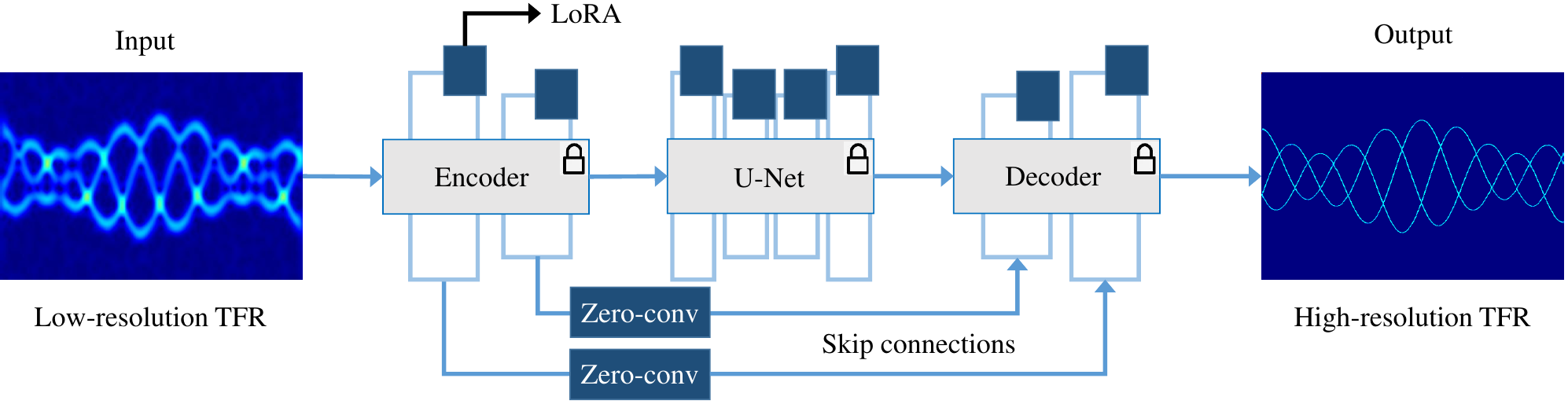}}
\caption{\small Overview of our proposed LoRA-SD for TFR super-resolution. We firstly employ pre-trained SD-Turbo \citep{sauer2024fast} as basic mapping function from input TFR with low-resolution to output TFR with high-resolution. Then we fix parameters of SD-Turbo and use LoRA \citep{hu2021lora} in each module of SD-Turbo including encoder, U-Net, and decoder. Additionally, we introduce skip connections using Zero-convs \citep{Zhang_2023_ICCV} between encoder and decoder. Among LoRA-SD, parameters denoted by blue boxes are trainable using adversarial optimization.}
\label{fig:lorasd}
\vspace{-3mm}
\end{figure*}

\textbf{TFA-based ISAR imaging for rigid-body targets} seeks to generate time-frequency representations (TFRs) with high resolution, and existing methods can be broadly categorized into two groups. \textbf{Linear TFA methods}, such as short-time Fourier transform (STFT) \citep{Chen2017} and Wigner-Ville distribution (WVD) \citep{Ren2016}, are computationally efficient but constrained by the uncertainty principle \citep{Cohen1995}, which limits simultaneous improvements in both time and frequency resolutions. \textbf{Advanced linear methods}, like sparse Bayesian learning (SBL) \citep{Hong2019} and fast mean field (MF) \citep{Wang2019}, have been introduced for better TFR quality. However, their performance heavily depends on the robustness of time-invariant coefficient estimation. \textbf{Nonlinear TFA methods}, such as those based on conditional generative adversarial networks (CGAN) \citep{Wehner2021} and U-Net architectures \citep{Pan2022,yang2022micro}, have shown promise in enhancing TFRs. Nonetheless, these methods often face challenges in generalization, particularly when significant distribution gaps exist between training and test radar datasets \citep{Dai2023}.


To overcome the limitations of existing methods, we propose a novel TFA-based ISAR imaging method, called \textbf{low-rank adaptation of the pre-trained stable diffusion (LoRA-SD)}. This approach leverages the pre-trained SD-Turbo \citep{sauer2024fast}, a highly regarded large visual model, alongside LoRA \citep{hu2021lora}, a popular fine-tuning technique for large models. Additionally, we incorporate adversarial training to further refine LoRA-based fine-tuning. Our approach addresses the drawbacks of existing methods through the following innovations: 1)	\textbf{High-Resolution TFR as Curve Refinement}. TFA aims to enhance TFR resolution from low-resolution counterparts, analogous to pixel-to-pixel curve refinement. Large visual models like SD-Turbo \citep{sauer2024fast} excel in perceiving texture features, including curves as a special case, making them ideal for this task. 2) \textbf{Robust Generalization via LoRA}. The distribution gap between the visual data used for pre-training SD-Turbo and the radar data used for TFA-based ISAR imaging can negatively impact performance. LoRA effectively mitigates this gap, ensuring robust generalization. 3) \textbf{Unrestricted by the Uncertainty Principle}. Unlike traditional TFA methods, our approach is not bound by the uncertainty principle, allowing simultaneous improvements in time and frequency resolutions.


\section{Methodology}

\subsection{Time-Frequency Analysis Method}

The radar signal in the azimuth domain at a specific range cell for a target with multiple scatterers can be modeled as a linear combination of multiple linear frequency modulation (LFM) components:
\begin{align}
\begin{split}
    s(t)=\sum_{n=1}^N a_n {\rm{exp}} (j 2\pi f_n t + j \pi k_n t^2)
\end{split}
\end{align}
where $N$ represents the number of scatterers, $a_n$ denotes the complex amplitude of the $n$-th scatter, $f_n$ and $k_n$ are the center frequency and chirp rate of the $n$-th scattering point, respectively.
To estimate the time-varying Doppler frequencies, the TFR of the signal can be obtained using the STFT:
\begin{align}
\begin{split}
    {\rm{STFT}}(t,w)=\int_{-\infty}^{+\infty} s(\tau) h(\tau-t) {\rm{exp}}(-jw \tau) d \tau
\end{split}
\end{align}
where $h(\cdot
)$ is the window function. The resulting spectrum is given by $S(t,w)=|{\rm{STFT}}(t,w)|^2$. 
By substituting spectrums (TFR) $\boldsymbol{S}=\{S(t,w)\}_{t=1,w=1}^{t=N,w=M}$ into the flowchart of Fig. \ref{fig:model}, we have a series of target images at different slow times, where $N$ and $M$ are separately the number of discrete time and frequency. 
Due to the tradeoff between time and frequency resolution inherent in conventional TFR methods, the resolution of TFR $S(t,w)$ is often limited.
To address this limitation, we enhance the low-resolution TFR using our proposed LoRA-SD model, which achieves super-resolution and improves the quality of the resulting ISAR images through the range-instantaneous Doppler (RID) algorithm.
The details of LoRA-SD are depicted in the following.

\subsection{Time-Frequency Representation Super-Resolution}

We introduce a novel approach, termed LoRA-SD, to achieve super-resolution in TFRs, specifically designed for high-resolution ISAR imaging of rigid-body targets. The architecture, illustrated in Fig. \ref{fig:lorasd}, consists of two main components: a pre-trained large visual model (SD-Turbo) \citep{sauer2024fast} and LoRA adapters \citep{hu2021lora} for fine-tuning.


\noindent \textbf{Pre-trained SD-Turbo:}
Observing that curve features in TFRs are a specific and relatively simple form of texture, we utilize the robust texture representation capabilities of SD-Turbo, a pre-trained large visual model, for TFR super-resolution.
The pre-trained encoder, U-Net, and decoder in SD-Turbo as shown in Fig. \ref{fig:lorasd} serve as the backbone for mapping low-resolution TFRs to high-resolution counterparts and can be written by
$\hat{\boldsymbol{S}}=\boldsymbol{G}_{\boldsymbol{\theta}}(\boldsymbol{S})$, where $\boldsymbol{G}_{\boldsymbol{\theta}}$ denotes pre-trained SD-Turbo, where the detailed architecture can be found in \citep{sauer2024fast}.


\noindent \textbf{LoRA adapters:}
To mitigate the distribution gap between optical image data used in pre-training and radar data used for TFR super-resolution, we introduce LoRA adapters \citep{hu2021lora} into the convolutional and transformer layers of SD-Turbo as shown in Fig. \ref{fig:lorasd}. 
We constrain the update of a pre-trained weight matrix $\boldsymbol{W}_0 \in \mathbb{R}^{d \times k}$ in SD-Turbo by representing its update with a low-rank decomposition $\boldsymbol{W}_0+\Delta \boldsymbol{W}=\boldsymbol{W}_0+\boldsymbol{B}\boldsymbol{A}$, where $\boldsymbol{B} \in \mathbb{R}^{d \times r}$, $\boldsymbol{A} \in \mathbb{R}^{r \times k}$, and we have ${\rm{rank}} \ r \ll {\rm{min}}(d,k)$. \textbf{This approach effectively reduces trainable parameters while ensuring efficient fine-tuning}. During training, $\boldsymbol{W}_0$ is frozen while $\boldsymbol{A}$ and $\boldsymbol{B}$ contain trainable parameters. 
We modify the forward pass by:
\begin{align}
\begin{split}
    \boldsymbol{h}=\boldsymbol{W}_0 \boldsymbol{x} + \Delta \boldsymbol{W} \boldsymbol{x}=\boldsymbol{W}_0 \boldsymbol{x}+\boldsymbol{B}\boldsymbol{A}\boldsymbol{x}
\end{split}
\end{align}

\noindent \textbf{Zero conv:}
We use zero conv introduced by \cite{Zhang_2023_ICCV} to maintain consistency between input and output. Zero conv is a $1 \ \times \ 1$ convolution with both weight and bias initialized to zero. As shown in Fig. \ref{fig:lorasd},
we employ zero conv as soft skip connections between encoder and decoder blocks of SD-Turbo \citep{sauer2024fast}.
Specifically, there are 4 encoder blocks and 4 decoder blocks. \textbf{We use zero conv to connect the $i$-th encoder block and the $(4-i+1)$-th decoder block for $i=1,...,4$}.

For simplicity and consistency, we denote the SD-Turbo equipped with LoRA adapters and zero conv as $\boldsymbol{G}_{\boldsymbol{\theta},\boldsymbol{\phi}}(\cdot)$, where $\boldsymbol{\phi}$ is parameters of LoRA and zero conv.

\begin{table}[!t]
  \caption{\small Motion parameters of simulated rigid-body target.}
  \vspace{-3mm}
\begin{center}
  \resizebox{0.45\textwidth}{!}{
  \begin{tabular}{c|ccc|c}
    \toprule
    {{Motion}} & {}
    & {Frequency (Hz)} & {} & {{No.}} \\
    \midrule 
    Spin & 3.0:0.1:4.0 & -  & - & 396 \\
    Precession & 1.5:0.2:2.5 & 1.5:0.2:2.3  & - & 1980 \\
    Nutation & 1.6:0.2:2.0 & 1.6:0.2:2.0 & 0.8:0.2:1.2 & 1782 \\  \bottomrule
  \end{tabular}}
  \end{center}
  \vspace{-2mm}
  \label{table1}
   \vspace{-2mm}
\end{table}

\begin{figure}[!t]
        \centering
        \centerline{\includegraphics[width=8.5cm]{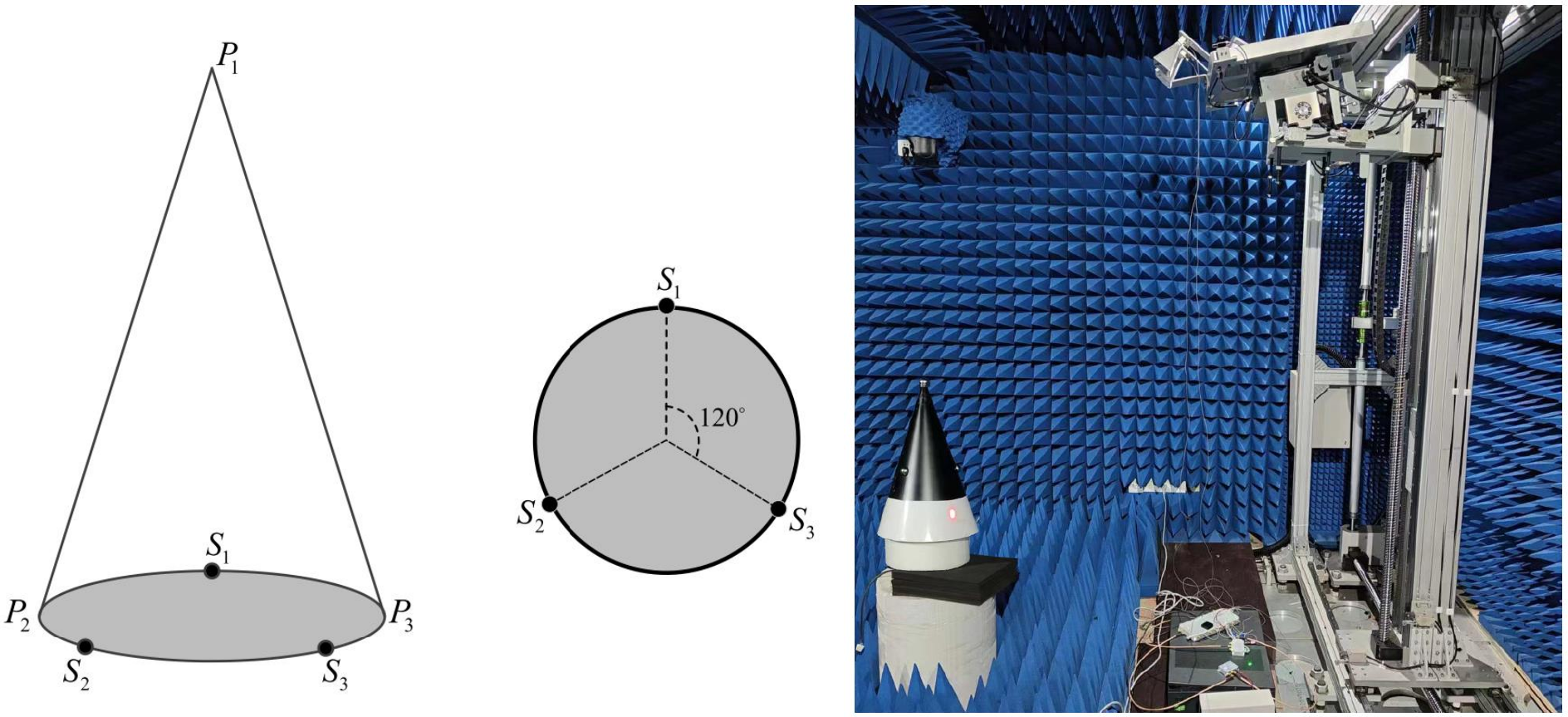}}
    \caption{\small \textbf{Left:} The model of rigid-body target. $\{S_i\}_{i=1}^3$ are common scattering centers and their positions are fixed on the surface of the cone. $\{P_j\}_{j=1}^3$ are scattering centers related to the geometric structure. Among them, $P_1$ is a common scattering center. $P_2$ and $P_3$ are equivalent scattering centers, and their positions are jointly determined by the direction between LOS of the radar and the cone's central axis. \textbf{Right:} Scenario for recording the measured radar echos.} 
    \label{fig:data}
    \vspace{-3mm}
\end{figure}

\subsection{Model Training and Inference}

At training time, we aim to optimize LoRA-SD by obtaining output TFRs with high-resolution given the corresponding input TFRs with low-resolution. Inspired by CycleGAN formulation \citep{zhu2017unpaired} for paired pixel2pixel translation, we propose to optimize our model by playing min-max game using the following objective w.r.t. trainable parameters of $\boldsymbol{G}_{\boldsymbol{\phi}}$ and $\boldsymbol{D}_{\boldsymbol{\Phi}}$ as ${\rm{\mathop{\arg\min}\limits_{\boldsymbol{G}_{\boldsymbol{\phi}}}}} \mathcal{L}$ and ${\rm{\mathop{\arg\max}\limits_{\boldsymbol{D}_{\boldsymbol{\Phi}}}}} \mathcal{L}$ iteratively, we have:
\begin{align} \label{eq:adv}
\begin{split}
\mathcal{L}& = \mathbb{E}_{\boldsymbol{S},\boldsymbol{Q}} [\|\boldsymbol{G}_{\boldsymbol{\phi}}(\boldsymbol{S})-\boldsymbol{Q}\|_2^2] \\
& + \alpha \mathbb{E}_{\boldsymbol{Q}}[{\rm{log}} \boldsymbol{D}_{\boldsymbol{\Phi}}(\boldsymbol{Q})]+\beta \mathbb{E}_{\boldsymbol{S}}[{\rm{log}}(1-\boldsymbol{D}_{\boldsymbol{\Phi}}(\boldsymbol{G}_{\boldsymbol{\phi}}(\boldsymbol{S})))]
\end{split}
\end{align} 
where $\boldsymbol{D}_{\boldsymbol{\Phi}}$ is an adversarial discriminator that aims to classify high-resolution TFRs from the corresponding low-resolution TFRs. We use the CLIP model as a backbone following \citep{kumari2022ensembling} \textbf{and only fine-tune the last fully-connection layer of $\boldsymbol{D}_{\boldsymbol{\Phi}}$}. $\boldsymbol{S}$ and $\boldsymbol{Q}$ separately represent low-resolution TFRs and their high-resolution counterparts.

During inference, we first use the well fine-tuned LoRA-SD to obtain high-resolution TFRs by $\boldsymbol{G}_{\boldsymbol{\theta},\boldsymbol{\phi}}(\boldsymbol{S})$, then we substitute the TFRs into the flowchart of Fig. \ref{fig:model}, the super-resolution ISAR imaging can be obtained by RID algorithm.

\begin{figure}[!t]
        \centering
        \centerline{\includegraphics[width=8.5cm]{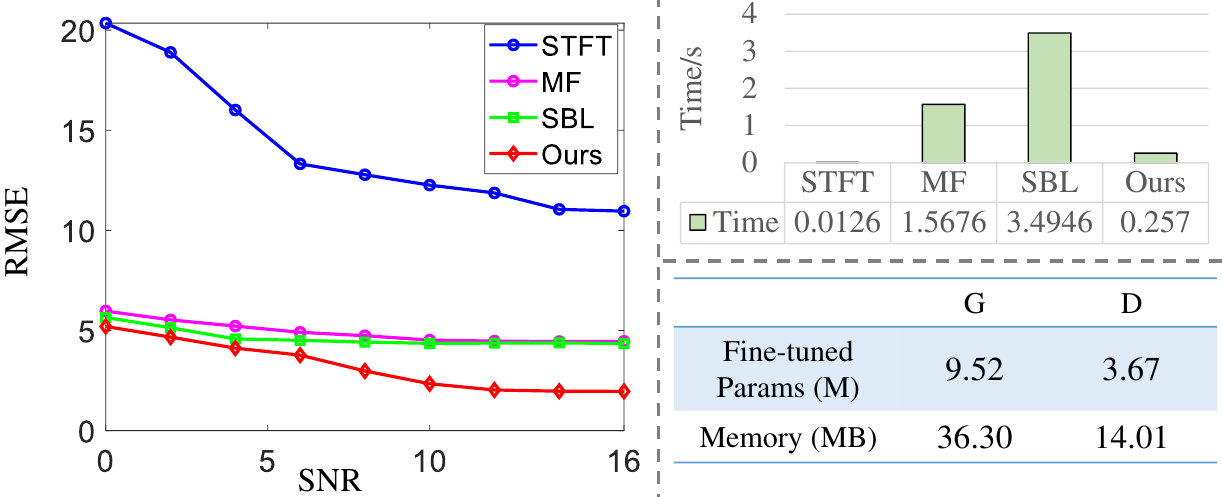}}
    \caption{\small \textbf{Left} The frequency estimation error evaluated on the simulated radar data. \textbf{Top right:} The averaged inference time per data. \textbf{Bottom right:} Trainable parameters and memory usage of LoRA-SD.}
    \label{fig:sim-xx}
    \vspace{-3mm}
\end{figure}

\section{Experiments}

\subsection{Experimental Setups}

\textbf{1) Simulated Radar Data:}
We simulate radar echoes from a rigid-body target exhibiting various micro-motions. The target's geometry and motions are detailed in Table. \ref{table1} and Fig. \ref{fig:data} (left).
Radar system parameters include a carrier frequency of 10 GHz, bandwidth of 400 MHz, and pulse repetition frequency of 1 kHz. The target has a height of 195 cm and a bottom surface radius of 51.6 cm.

\textbf{2) Measured Radar Data:}
The height of the meta conical target is $65$ cm, and the radius of the bottom surface is $17.2$ cm. At the height of $30$ cm from the bottom, there are three meta semispheres with a diameter of $3.8$ cm. The positions of the three semispheres on the cone are equidistant. The conical target is placed on an electric rotating platform, and its central axis coincides with the rotation axis of the rotating platform. The carrier frequency is 10 GHz, the bandwidth is 1 GHz, and the range resolution is 15 cm. The transmitting and receiving antennas of the radar are placed on a platform with an adjustable pitch angle, as shown in Fig. \ref{fig:data} (right).
 
\textbf{2) Implementation Details:}
Radar echoes from three motions are mixed and randomly split into training and test datasets with an 8:2 ratio. We use pre-trained SD-Turbo and only fine-tune parameters in LoRA, zero conv, and fully-connected layer of discriminator.
The model is trained using AdamW \citep{loshchilov2017decoupled} optimizer for 100 epochs with a learning rate of $5\times 10^{-6}$ and a batch size of 2. $\alpha=\beta=0.5$. We update the trainable parameters in $\boldsymbol{G}_{\boldsymbol{\phi}}$ and $\boldsymbol{D}_{\boldsymbol{\Phi}}$ iteratively. All experiments are conducted on an NVIDIA A6000 GPU.

\textbf{3) Baselines and Metrics:}
We compare our model with STFT \cite{Chen2017}
, SBL \cite{Hong2019}, and MF \cite{Wang2019}.
The Root Mean Square Error (RMSE) is used to evaluate frequency estimation quality under different Signal-Noise Ratios (SNRs) \cite{Dai2023}.

\begin{figure*}[!t]
        \centering
        \centerline{\includegraphics[width=16cm]{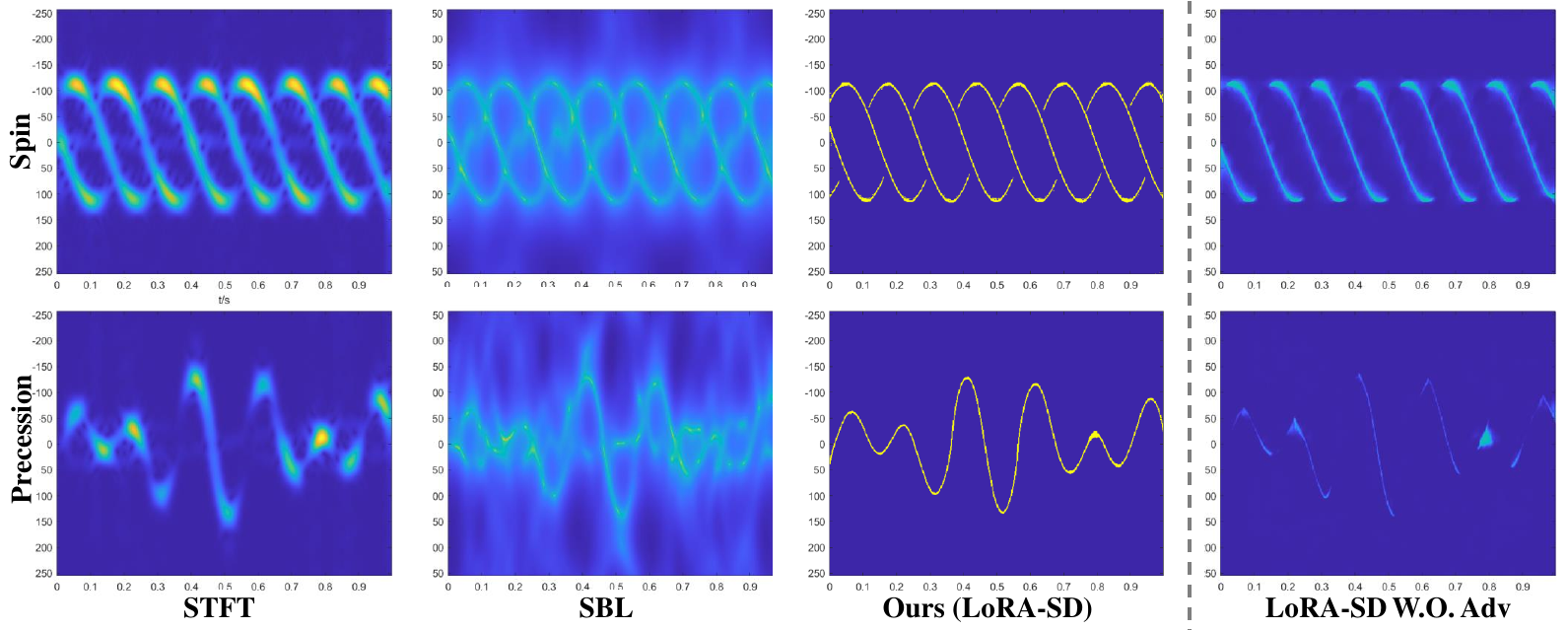}}
        \vspace{-1mm}
    \caption{\small \textbf{Left} TFRs under spin and precession on the measured radar data when SNR=5dB. \textbf{Right} TFRs of our model W.O adversarial training.}
    \label{fig:sim}
    \vspace{-3mm}
\end{figure*}

\begin{figure*}[t]
\centering
\centerline{\includegraphics[width=16cm]{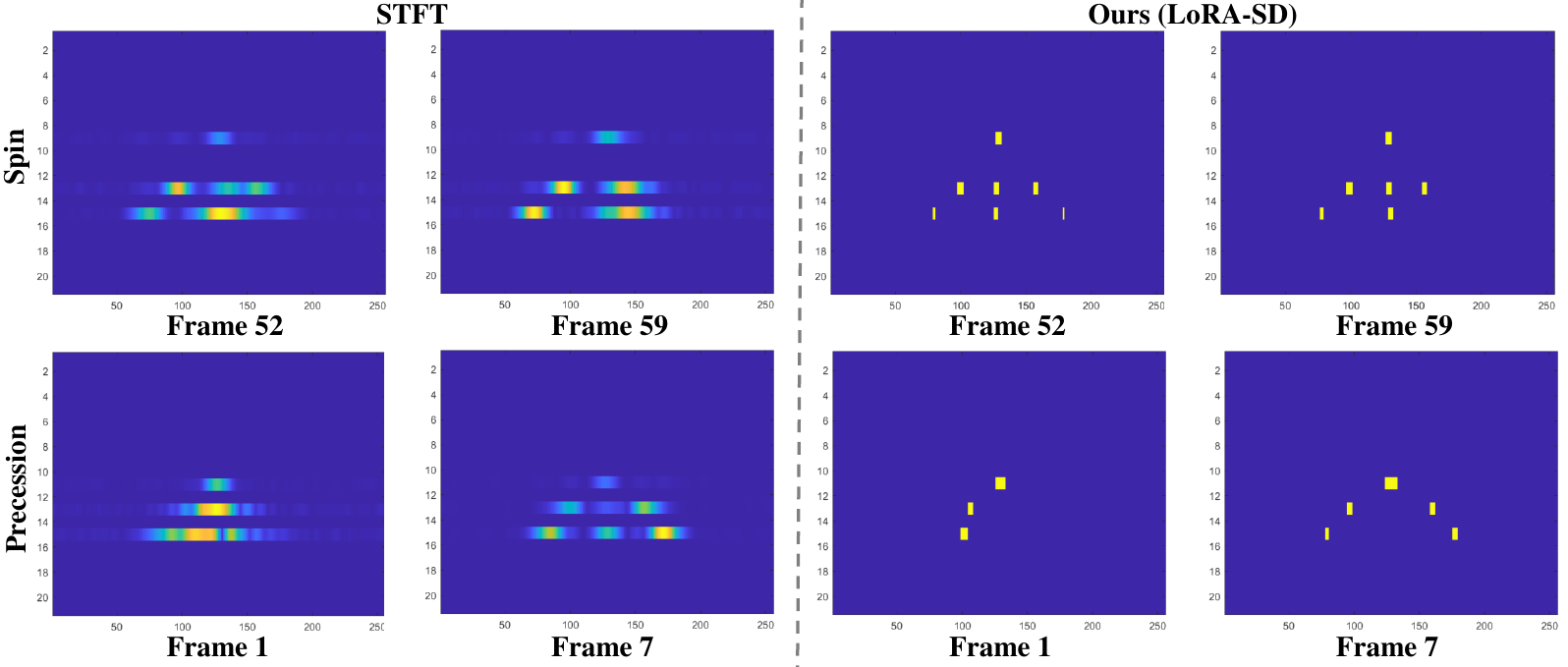}}
\vspace{-1mm}
\caption{\small ISAR imaging results for rigid-body target with different frames using the measured radar data under spin and precession.}
\label{fig:isar}
\vspace{-3mm}
\end{figure*}


\subsection{Quantitative Results on the Simulated Dataset}
Numerical experiments are reported in Fig. \ref{fig:sim-xx} on the simulated radar data. We drawn three observations as follows.

\textbf{1):} As shown in Fig. \ref{fig:sim-xx} left, LoRA-SD 
achieves significantly lower frequency estimation errors compared to STFT, SBL, and MF across various SNRs ranging from 0dB to 16dB with interval 2dB. This demonstrates the robustness of our model against noise due to its superior perception of texture features, especially for curves. On the contrary, SBL and MF are developed for computing TFR by solving TVAR model \citep{mukhopadhyay1997parametric}, whose performance is sensitive to selecting hyper-parameters.

\textbf{2):} As shown in Fig. \ref{fig:sim-xx} top right, regarding computational efficiency, LoRA-SD offers promising real-time performance, striking a balance between effectiveness and efficiency.

\textbf{3):} As shown in Fig. \ref{fig:sim-xx} bottom right, the memory usage of LoRA-SD is acceptable due to its limited trainable parameters.

\subsection{Qualitative Results on the Measured Dataset}
We perform experiments on the measured dataset and report the results of our model and competitors in Fig. \ref{fig:sim} and Fig. \ref{fig:isar}. Several observations can be obtained as follows.

\textbf{1):} According to Fig. \ref{fig:sim} left, 
our LoRA-SD model achieves high-resolution TFRs despite being fine-tuned only on simulated radar data with a different bandwidth. This showcases its strong generalization capability. 

\textbf{2):} As shown in Fig. \ref{fig:sim} right, TFRs calculated by LoRA-SD without adversarial training are inaccurate compared with their LoRA-SD counterparts. It suggests the importance of our adversarial-based objective in Eq. \ref{eq:adv}.

\textbf{3):} 
We insert LoRA-SD into the RID algorithm \citep{Chen2016}, as shown in Fig. \ref{fig:model}, and report the ISAR images of STFT and LoRA-SD for rigid-body target under different motion at various frames in Fig. \ref{fig:isar}. 
Our model outperforms STFT, providing sharply focused images where scatterers can be distinctly counted. Additionally, scatterer occlusion and reappearance effects are accurately captured, indicating the potential of LoRA-SD in 3D pose estimation for rigid-body targets.

\section{Conclusion}
We propose a novel TFR super-resolution method called LoRA-SD for high-resolution ISAR imaging for rigid-body target with complex motions like spin and precession. Generalization capability is verified by the measured radar data.

\small
\bibliographystyle{IEEEtranN}
\bibliography{references}

\end{document}